\documentclass[final]{cvpr}
\usepackage{times}
\usepackage{epsfig}
\usepackage{graphicx}
\usepackage{amssymb}
\usepackage{makecell}

\usepackage{amsmath}
\usepackage[pagebackref=true,breaklinks=true,colorlinks,bookmarks=false]{hyperref}

\def\cvprPaperID{74} % *** Enter the CVPR Paper ID here
\def\confYear{CVPR 2021}

\usepackage{multirow}
\usepackage{subcaption}

\begin{document}

% }
% \title{One-shot action recognition towards novel assistive therapies}
\title{One-shot action recognition in challenging therapy scenarios}

\author{Alberto Sabater $^{1,2}$ ~~~ Laura Santos $^{1,4}$ ~~~
José Santos-Victor$^{1}$ ~~~ Alexandre Bernardino$^{1}$
\vspace{0.1cm} \\ Luis Montesano$^{2,3}$ ~~~ Ana C.~Murillo$^{2}$\\
~ \vspace{-0.3cm} \\
{\small $^{1}$ISR-Lisboa, Instituto Superior T\'ecnico, Universidade de Lisboa, Portugal},\\
{\small $^{2}$DIIS-I3A, Universidad de Zaragoza, $^{3}$Bitbrain Technologies, Spain}\\
{\small $^{4}$Politecnico di Milano, Italy}
}

\maketitle
% \thispagestyle{empty}
% \pagestyle{empty}

%%%%%%%%%%%%%%%%%%%%%%%%%%%%%%%%%%%%%%%%%%%%%%%%%%%%%%%%%%%%%%%%%%%%%%%%%%%%%%%%
\begin{abstract}

One-shot action recognition aims to recognize new action categories from a single reference example, typically referred to as the anchor example. This work presents a novel approach for one-shot action recognition in the wild that computes motion representations robust to variable kinematic conditions. One-shot action recognition is then performed by evaluating anchor and target motion representations. We also develop a set of complementary steps that boost the action recognition performance in the most challenging scenarios. Our approach is evaluated on the public NTU-120 one-shot action recognition benchmark, outperforming previous action recognition models.  Besides, we evaluate our framework on a real use-case of therapy with autistic people. These recordings are particularly challenging due to high-level artifacts from the patient motion. Our results provide not only quantitative but also online qualitative measures, essential for the patient evaluation and monitoring during the actual therapy.
% \footnote{This research has been funded by FEDER/Ministerio de Ciencia, Innovación y Universidades/Agencia Estatal de Investigación RTC-2017-6421-7
% and PGC2018-098817-A-I00, DGA T45 17R/FSE, the Office of Naval Research Global project ONRG-NICOP-N62909-19-1-2027, and the Funda\c{c}\~ao para a Ci\^encia e Tecnologia (FCT) project UIDB/50009/2020 and PhD scholarship SFRH/BD/145040/2019.}

\end{abstract}

%%%%%%%%%%%%%%%%%%%%%%%%%%%%%%%%%%%%%%%%%%%%%%%%%%%%%%%%%%%%%%%%%%%%%%%%%%%%%%%%
%%%%%%%%%%%%%%%%%%%%%%%%%%%%%%%%%%%%%%%%%%%%%%%%%%%%%%%%%%%%%%%%%%%%%%%%%%%%%%%%
\section{Introduction}

Human action recognition is a challenging problem with high relevance for many application fields such as human-computer interaction, virtual reality or medical therapy analysis.
Certain works use raw appearance information for the action recognition tasks \cite{donahue2015long, zhang2012spatio}.
Most recent works use skeleton-based representations \cite{zhang2019view, liu2020disentangling}. These representations encode the human pose independently of appearance, surroundings and being more robust to occlusions.

\begin{figure}[!t]
\centering
\includegraphics[width=0.99\linewidth]{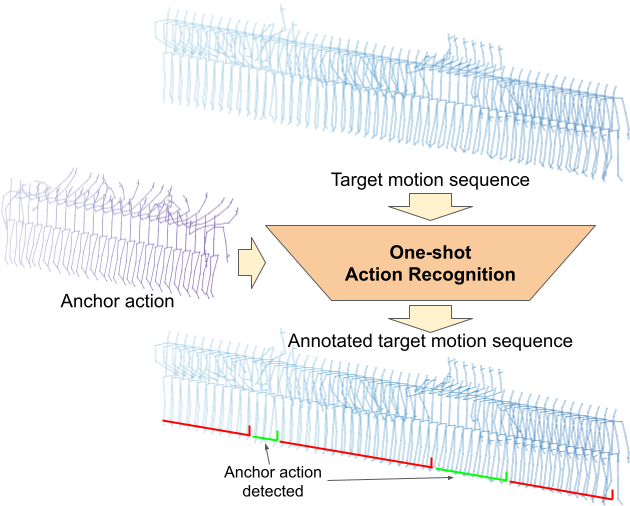}
\caption{Overview of the proposed action recognition method. An anchor action is processed and compared against a target motion sequence. Output results show when the anchor action has been performed within the target motion sequence. 
}
\label{fig:det_overview}
\end{figure}

Real world scenarios often require the capability to recognize 
new action categories 
that cannot be learned, due to their creation on the fly or data limitations.
Action classifiers that handle this problem \cite{vinyals2016matching, liu2017skeleton} of learning from limited data are based on encoding motion sequences into meaningful descriptors. Then, to recognize a new target sequence, they evaluate the similarity of the target descriptor with the descriptors from one (one-shot) or few (few-shot) anchor labeled actions. 
Learning discriminative action encodings and their application in real scenarios are still open challenges. This is partially due to variable motion recording set-ups and unconstrained action execution rules (unsegmented action sequences, multiple consecutive executions, heterogeneous action executions, etc.).

This work tackles the problem of 
one-shot action recognition in unconstrained motion sequences.
Our novel skeleton-based solution\footnote{Code, learned models and supplementary video can be found in: https://sites.google.com/a/unizar.es/filovi/}
is summarized in Fig. \ref{fig:det_overview}.
In the proposed workflow, a
stream of skeleton poses is encoded in an online manner, generating a descriptor at each time-step.
Comparing the descriptors from a given reference anchor action with descriptors from 
a target video, we can detect when the anchor action has been performed within the target sequence.

The main components of our work are two fold: 
1) a motion encoder, based on geometric information, which is robust to 
heterogeneous movement kinematics;
2) the actual one-shot action recognition step, based on evaluating the similarity between an anchor action and target motion encodings.
We propose a set of improvements to this final action recognition step designed to achieve a more accurate action recognition in the wild. These improvements include an extended anchor action representation and a dynamic threshold that discriminates challenging action sequences. Besides, the proposed action recognition approach can be easily extended to a few-shot problem, if multiple anchor actions are available.

The presented approach is validated on a generic and public one-shot action recognition benchmark, the NTU RGB+D 120 dataset \cite{liu2019ntu}, where it outperforms the existing baseline results. 
Besides, we exhaustively analyze the performance of our system on data from a real application, automatic analysis of therapies with autistic people (based on gesture and action imitation games). Evaluation shows our proposed improvements to work in the wild, managing to overcome challenging motion artifacts that are particular for this real environment. Final outcome provides online and real-time quantitative and qualitative results, essential to evaluate the patient attention and coordination.

%%%%%%%%%%%%%%%%%%%%%%%%%%%%%%%%%%%%%%%%%%%%%%%%%%%%%%%%%%%%%%%%%%%%%%%%%%%%%%%%
%%%%%%%%%%%%%%%%%%%%%%%%%%%%%%%%%%%%%%%%%%%%%%%%%%%%%%%%%%%%%%%%%%%%%%%%%%%%%%%%
%%%%%%%%%%%%%%%%%%%%%%%%%%%%%%%%%%%%%%%%%%%%%%%%%%%%%%%%%%%%%%%%%%%%%%%%%%%%%%%%
%%%%%%%%%%%%%%%%%%%%%%%%%%%%%%%%%%%%%%%%%%%%%%%%%%%%%%%%%%%%%%%%%%%%%%%%%%%%%%%%
\section{Related Work}

This section summarizes the most relevant contributions for skeleton-based action recognition, making special emphasis on N-shot action recognition methods.

\subsection{Skeleton representations for action recognition}
Although many skeleton-based action recognition approaches use the raw Cartesian pose coordinates with no special pre-processing \cite{perez2021interaction, liu2019ntu}, other works research on skeleton data transformations to achieve view-invariant coordinates or compute new geometric features. 
The approach in \cite{zhang2019view} trains a Variational Autoencoder to estimate the most suitable observation view-points, and transforms the skeletons to those view-points. Other approach applies  
view-invariant transformations to the skeleton coordinates \cite{su2020predict}. Regarding the computation of new skeleton features, earlier work \cite{chen2010learning} computes multiple geometric features including joint distances and orientation, distances between joints, lines and planes, velocity and acceleration. More recent approaches describe a set of geometric features based on distances between joints and lines \cite{zhang2017geometric} or propose to use pair-wise euclidean joint distances and two-scale motion speeds~\cite{yang2019make}. 

Our approach uses a skeleton representation that combines a view-invariant transformation of the skeleton Cartesian coordinates with the computation of a set of additional geometric features.

\subsection{Skeleton-based neural networks for action recognition}
Recurrent neural networks have been widely applied to learn temporal dependencies for action recognition approaches. 
\cite{su2020predict} uses a Variational Autoencoder with bidirectional GRU layers for unsupervised training,
and work in \cite{liu2017global} presents a recurrent attention mechanism that improves iteratively. 
Other approaches rely on 1D Convolutional Neural Networks (CNN), such as 
\cite{yang2019make}, that uses a ResNet backbone, and
\cite{zhang2019view} in which a CNN is combined with LSTM networks.
Another common architecture within action recognition approaches is Graph Convolutional Networks, such as the models proposed in \cite{liu2020disentangling} and \cite{cheng2020skeleton}, that reduce the computational complexity with flexible receptive fields.

In our method, a Temporal Convolutional Network (TCN) (\cite{bai2018empirical, oord2016wavenet}) is chosen to encode temporal action segments into fixed-length representations. TCNs have already shown a good action recognition performance, easing the interpretability of their results \cite{kim2017interpretable}. TCNs use one-dimensional dilated convolutions to learn long-term dependencies from variable length input sequences. Convolutions allow parallelizing computations allowing fast inference and performing equally or even better than RNNs in sequence modeling tasks by exhibiting longer memory \cite{bai2018empirical}.

\subsection{N-shot action recognition}
N-shot action recognition is still an active area of research. We find earlier methods like 
\cite{fanello2013one} and \cite{konevcny2014one} that use HOF and HOG features for one-shot action recognition in RGB+D videos.
More recently, \cite{liu2019ntu} and \cite{liu2017skeleton} use a 2D Spatio-Temporal LSTM Network, the latter with a bidirectional pass. 
Other applications join visual and semantic information to perform zero-shot action recognition, 
such as \cite{hahn2019action2vec}, that uses hierarchical LSTMs and word2vec \cite{mikolov2013distributed},
and \cite{jasani2019skeleton}, that uses a Relation Network, a Spatial Temporal Graph Convolution Network (ST-GCN) and sent2vec. \cite{pagliardini2017unsupervised}.

Different from these methods, we propose the simple yet effective TCN described above as a feature extractor for one-shot action recognition. Additionally, we show how 
a robust variation of the approach
can boost the final recognition performance in challenging real world scenarios.

%%%%%%%%%%%%%%%%%%%%%%%%%%%%%%%%%%%%%%%%%%%%%%%%%%%%%%%%%%%%%%%%%%%%%%%%%%%%%%%%
%%%%%%%%%%%%%%%%%%%%%%%%%%%%%%%%%%%%%%%%%%%%%%%%%%%%%%%%%%%%%%%%%%%%%%%%%%%%%%%%
%%%%%%%%%%%%%%%%%%%%%%%%%%%%%%%%%%%%%%%%%%%%%%%%%%%%%%%%%%%%%%%%%%%%%%%%%%%%%%%%
\section{Proposed Framework}

\begin{figure}[!t]
\centering
\includegraphics[width=0.99\linewidth]{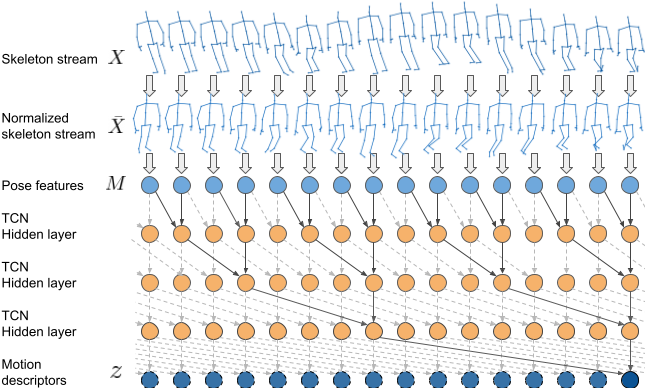}
\caption{
Motion descriptor generation. Input skeleton coordinates $X_n$ are normalized $\bar{X}_n$ and pose features $M$ are calculated. Pose features are processed by a TCN to generate motion descriptors $z$.
}
\label{fig:descr_gen}
\end{figure}

Figure \ref{fig:descr_gen} summarizes our motion description approach. It first normalizes input streams of skeleton data (computed using standard techniques \cite{zhang2012microsoft}), it generates sets of pose features that are encoded into motion descriptors by a TCN.
One-shot action recognition is performed evaluating the similarity between motion descriptors from anchor and target sequences.

\subsection{Pose normalization}
A human movement is defined by a set of $N$ poses $X = \{X_1,..,X_N\}$. 
Each pose $X_n$ is defined as a set of $J$ 3D body keypoint coordinates, $X_n = \{x_{1}^{n}, ..., x_{J}^{n}\}, x_{j}^{n} \epsilon \mathbb{R}^3$, composing what we name a skeleton.

\begin{figure}[!tb]
    \centering
    \includegraphics[width=0.7\linewidth]{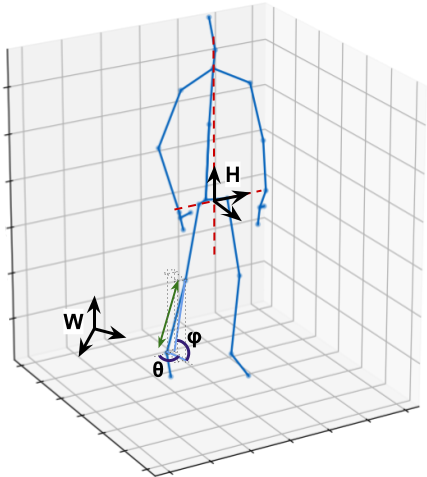}\\
    \caption{Skeleton representation. W and H refer to the original and transformed skeleton coordinate systems respectively. H axis are aligned with the vectors (dashed lines) that cross the human hip and spine. $\varphi$ and $\theta$ angles refer to the elevation and azimuth calculation in a bone from the leg.}
    \label{fig:skel_info}
\end{figure}

Human actions are frequently recorded in dynamic scenarios that involve different view-points and people moving and interacting freely. 
To achieve a better action recognition generalization, we normalize skeleton data by applying a per-frame coordinate transformation from the original coordinate system $W$ to a new one $H$, obtaining new \textbf{view} and \textbf{location-invariant} coordinate sets.
As represented in the Fig. \ref{fig:skel_info}, $H$ is set to have its origin at the middle point of the vector composed by the two hip keypoints. 
This hip vector is aligned to the new X axis and oriented to always have left and right hip X coordinates negative and positive values respectively. Similarly, the vector composed by the spine keypoint and the origin becomes the Y axis, leaving the Z axis to describe the depth information by being orthogonal to X-Y. 
The corresponding transformation $T^{HW}$ is applied to each 3D point in a pose $X_n$ to obtain the new set of 3D keypoint coordinates $\hat{X}_n$ as: 
\begin{eqnarray}
    \hat{X}_n = T_{n}^{HW} X_n
\end{eqnarray}

Regardless of the camera configuration, action sequences can be performed by different people with heterogeneous heights. To get \textbf{scale-invariant} coordinates $\bar{X}_n$, each skeleton $\hat{X}_n$ is scaled to a predefined size. In particular, since the joints defining the torso usually present little noise, we scale each skeleton to have a fixed torso length $\bar{L}$:
\begin{eqnarray}
    \bar{X}_n = \hat{X}_n * \frac{\bar{L}}{L_n}
\end{eqnarray}
\noindent where $L_n$ is the length of the corresponding torso and $\bar{L}$ is set as the average ratio between the torso length and the height (calculated from the NTU RBG+D 120 Dataset). 

\begin{figure}[!tb]
\centering
\begin{tabular}{c}
    \includegraphics[width=0.95\linewidth]{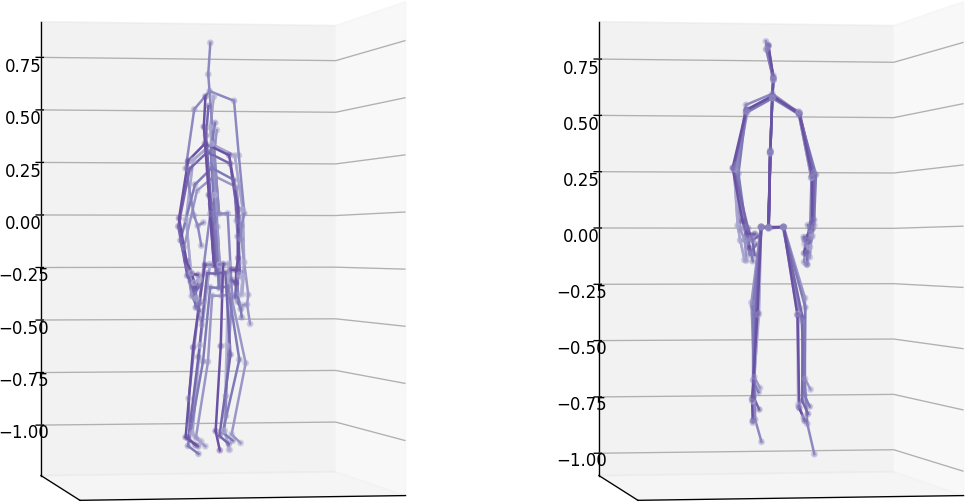}\\
    (a) skeletons from \textit{jump up} action\\
    \includegraphics[width=0.95\linewidth]{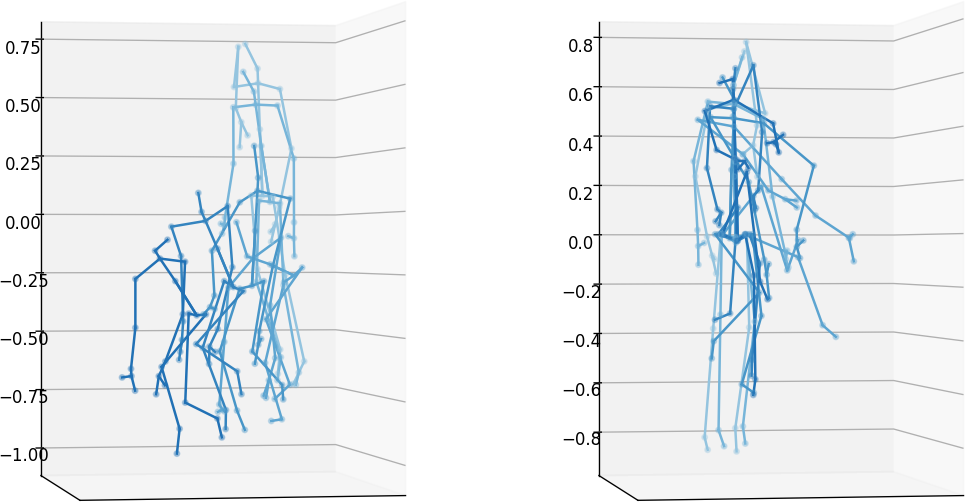}\\
    (b) skeletons from \textit{moving heavy objects} action
\end{tabular}
\caption{Pose normalization. Left plots show original skeleton coordinates of different actions from the NTU-120 RGB-D dataset. Right plots represent the same skeletons after applying the proposed pose normalization. 
}
\label{fig:skel_norm}
\end{figure}

Figure \ref{fig:skel_norm} shows original and normalized skeleton coordinates after applying the two proposed normalization steps.

\subsection{Pose Features}
\label{sec:feat_set}

The presented approach includes as final pose features the normalized coordinates $\bar{X}_n$, described above, and the following additional geometric features:
\begin{itemize}
    \item \textbf{Pair-wise keypoint distances $P_n$}, calculated as the Euclidean distance between each possible pair of skeleton joints, encode the pose relative to the $J$ skeleton joints. This set of features has the size of $\binom{J}{2}$.
    \item \textbf{\textit{Bone} angles $B_n$} from the original coordinates, calculated as the elevation $\varphi$ and azimuth $\theta$ (see Fig. \ref{fig:skel_info}) of 
    each vector composed by two connected joints.
    These angles encode the orientation of each bone relative to the world. This set of features has the size of $b \times 2$, being $b$ the number of bones within the skeleton.
\end{itemize}

\subsection{Motion descriptor generation from a TCN}

In order to generate motion representations based on the temporal context and not only static pose features,
we use a Temporal Convolutional Network (TCN) \cite{bai2018empirical, oord2016wavenet}. The TCN processes, as illustrated in Fig.~\ref{fig:descr_gen}, streams of pose features $M = \{\bar{X}, P, B\}$, and obtains motion embeddings $z$ (or descriptors).
This motion generation works in an online fashion, creating embeddings $z_{n} = TCN(M_{n-w:n})$ that encode, at the time $n$, all the motion from the last $w$ frames. This receptive field (memory) $w$ is implicitly defined by the TCN hyperparameters (details in Section \ref{sec:impl}).

\subsection{One-shot action recognition}

We formulate the one-shot action recognition as a simple similarity evaluation between the anchor embedding $z_a$ calculated to describe an anchor action and a stream of target embeddings $z_T$ extracted from a full video sequence.

For the \textbf{anchor action description}, we use the embedding associated to the last frame of the anchor action, i.e. $z_a$, assuming it encodes all the relevant previous motion information. Then, the \textbf{evaluation distance} at time $n$ is given by the distance between the anchor embedding and the target embedding $z_{T}(n)$ computed at time $n$: 
\begin{eqnarray}
    d_{1}(n) = D(z_{a},z_{T}(n))%,  \\ 
\end{eqnarray}

For the \textbf{embedding distance computation} $D$ we have explored several options. The cosine distance ($cos$) and the Jensen-Shannon divergence ($JS$) are the two best performing ones:
{\small
\begin{eqnarray}\label{eq:d1}
    D_{cos} ({z_1},{z_2})= 1 - \frac{{z_1} \cdot {z_2}}{  \|{z_1}\| \|{z_2}\|},  \\ %\nonumber
    D_{JS} ({z_1},{z_2})= \sqrt{\frac{KL(z_1 \left | \right | \bar{z}_{1,2}) + KL(z_2 \left | \right | \bar{z}_{1,2})}{2}}%, \\
\end{eqnarray}}%
\noindent where $z_1$ and $z_2$ are two motion embeddings, $KL$ is the \textit{Kullback-Leibler divergence} and $\bar{z}_{1,2}$ is the pointwise mean of $z_1$ and $z_2$.
Both functions are bounded between 0 and 1, being this last value the lowest similarity between two movement descriptors. 

The \textbf{final action recognition} is performed by thresholding the calculated distance. If $d_{1}(n)$ is below the acceptance threshold $\alpha$, we consider that the anchor action has been detected by frame $n$. This threshold value is set by evaluating the precision/recall curve over an evaluation dataset, as detailed in Section \ref{sec:val_ther_quant}.

\subsection{Improving action recognition in the wild}

Real action recognition applications (e.g. real medical therapies) involve artifacts that hinder the motion description and recognition. These issues are intensified in the one-shot recognition set-up, where the available labeled data is limited. In the following, we describe different improvements for the action recognition described previously, to get better performance in the wild.

\paragraph{Extended anchor action representation.}
Anchor actions are not only scarce but frequently also hard to consistently segment in a video sequence. Therefore, using just the last embedding generated for them can lead to noisy action descriptions. In order to get a better anchor action representation, we use a set of descriptors $z_A=\{z_{1}, ..., z_{m}\}$ composed by the ones generated at their last $m$ frames. 
The distance between a target embedding $z_{T}(n)$ and the anchor action is then set as the minimum distance to each element of the set of anchor embeddings ($z_A$):
    \begin{eqnarray}\label{eq:dm}
        d_{m}(n) = \min_{\forall z_{a} \in z_A} D(z_{a},z_{T}(n))
    \end{eqnarray}

\paragraph{Few-shot recognition.}
In case more than one anchor sequence is available, the set of anchor embeddings $z_A$ can be easily augmented with the embeddings generated for the different anchor sequences. 

\paragraph{Dynamic threshold.}
Most challenging scenarios can include actions consisting of maintaining a static pose or performing subtle movements. The descriptors generated from this type of actions are very similar to the descriptors from target idle poses (e.g. when no specific action is being performed).
If information of idle positions is available, we propose to use it to set a dynamic threshold that can better discriminate idle poses from actual actions. New threshold value is set to be the minimum between the original threshold $\alpha$ and the $10^{th}$ percentile $P_{10\%}$ of all the distances computed between a given idle target sub-sequence (identified within the target sequence) and the anchor action representation:

\begin{eqnarray}
    \alpha = \min ( \alpha, \mathop{P_{10\%}}_{n = a \ldots b} \{d(n)\}) ,
\end{eqnarray}
where $a$ and $b$ are the initial and final time steps of the idle interval, and $d$ is computed as in Eq. (\ref{eq:d1}) or Eq. (\ref{eq:dm}) depending on the anchor representation used.

%%%%%%%%%%%%%%%%%%%%%%%%%%%%%%%%%%%%%%%%%%%%%%%%%%%%%%%%%%%%%%%%%%%%%%%%%%%%%%%%
%%%%%%%%%%%%%%%%%%%%%%%%%%%%%%%%%%%%%%%%%%%%%%%%%%%%%%%%%%%%%%%%%%%%%%%%%%%%%%%%
%%%%%%%%%%%%%%%%%%%%%%%%%%%%%%%%%%%%%%%%%%%%%%%%%%%%%%%%%%%%%%%%%%%%%%%%%%%%%%%%
\section{Experiments}

This section details experimental validation of the proposed action recognition approach, including implementation and training details.

\subsection{Experimental setup}

\subsubsection{Datasets}\label{sec:datasets}

The proposed method is designed for one-shot online action recognition in real scenarios. In particular, our motivation is to automate the analysis of medical therapies. Since the available data in this setting is scarce, real therapy data is only used for evaluation, while a large public action dataset (NTU RGB+D 120) is used to learn the action encoding model as well as to validate the motion representation framework in a public benchmark.

\begin{figure}[!t]
\centering

\begin{subfigure}[b]{\linewidth}    % 0.49
    \includegraphics[width=0.32\linewidth]{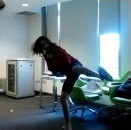}
    \includegraphics[width=0.32\linewidth]{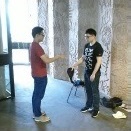}
    \includegraphics[width=0.32\linewidth]{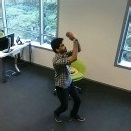}
    \caption{NTU RGB+D 120 dataset frames.}
    \label{fig:datasets_ntu}
\begin{subfigure}[b]{\linewidth}    % 0.49
    \includegraphics[width=0.32\linewidth]{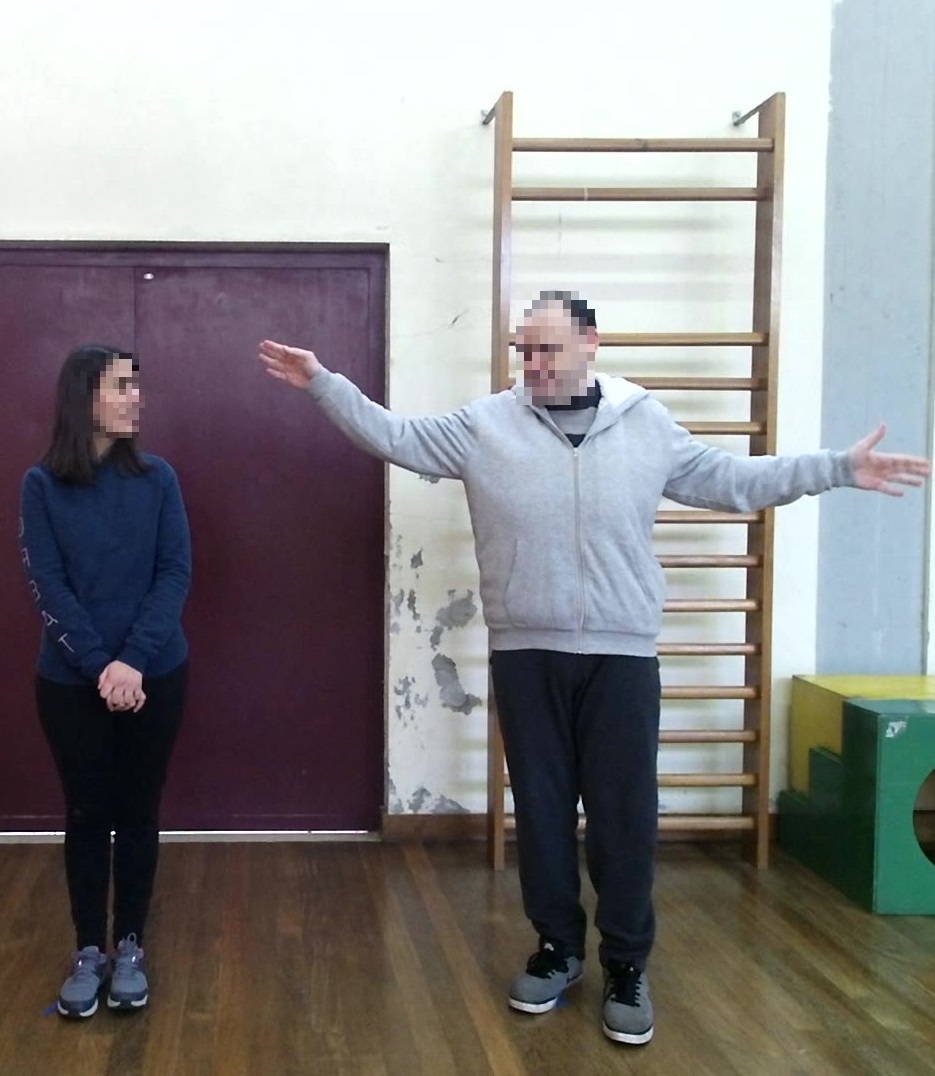}
    \includegraphics[width=0.32\linewidth]{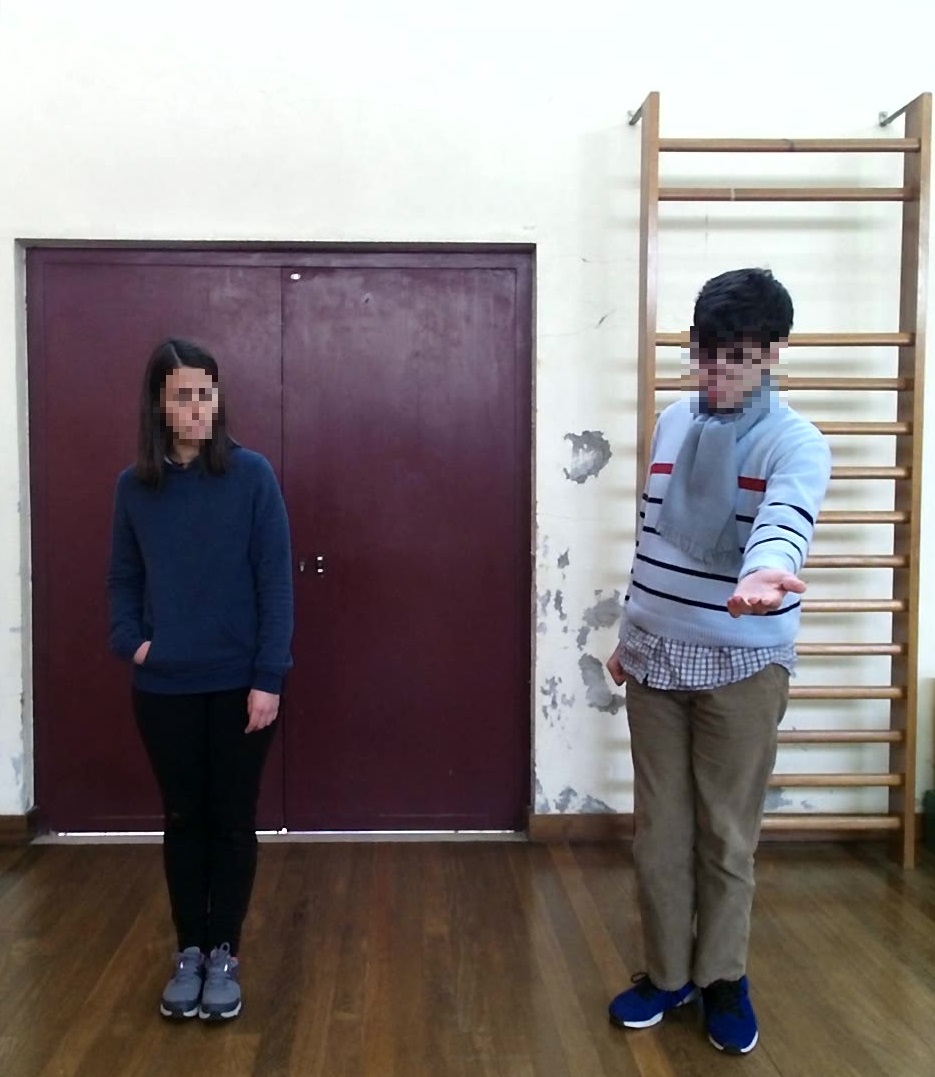}
    \includegraphics[width=0.32\linewidth]{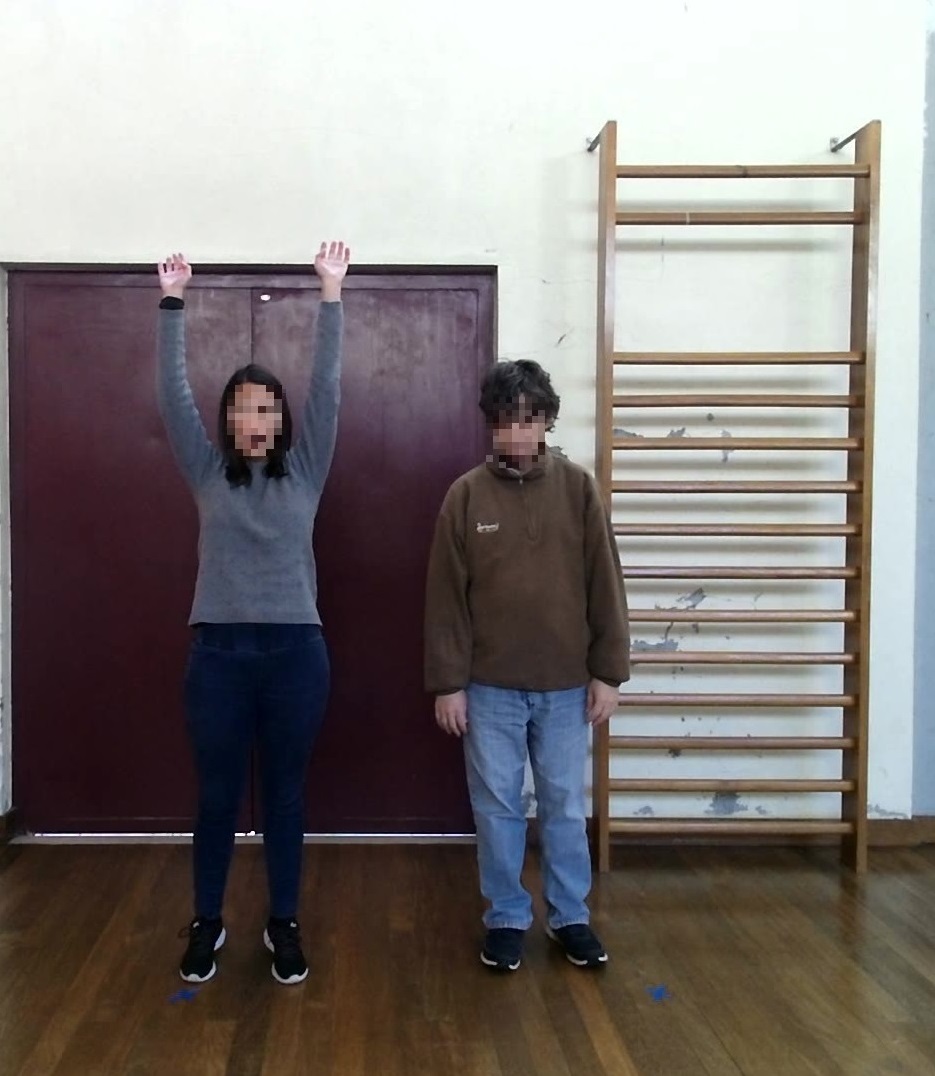}
    \caption{Therapy dataset frames.}
    \label{fig:datasets_ther}
\end{subfigure}
\end{subfigure}

\caption{
Sample frames from the training (a) dataset (\textit{kicking something}, \textit{handshaking} and \textit{shoot at the basket actions}) and evaluation (b) datasets (\textit{big}, \textit{give} and \textit{high gestures}).
}
\label{fig:datasets}
\end{figure}

% TRAINING DATASET
The \textbf{NTU RGB+D 120} dataset~\cite{liu2019ntu} (see Fig. \ref{fig:datasets_ntu}) is so far the largest dataset available for skeleton-based action classification. It contains 114,480 action samples belonging to 120 different categories with different complexities. Action sequences are recorded at 30 fps with different users, set-ups and points of view.

% EVAL DATASET
The \textbf{therapy dataset\footnote{http://vislab.isr.tecnico.ulisboa.pt/datasets/\#autism}} (see Fig. \ref{fig:datasets_ther}) has been acquired in real medical settings and consists of 57 different imitation games where the patient has to imitate the therapist \cite{Santos2019, santos2020interactive} action and gestures.
Each imitation game is composed of at least one anchor action and one patient imitation, but frequently more anchors (up to 3) or imitations appear in the games. Due to the nature of this data, patient motion is characterized by strong artifacts such as highly variable length imitations, uncommon resting poses and poor quality imitations.
Due to limited computational resources, sessions are recorded with a variable low frame rate of 12 fps on average.

\subsubsection{Implementation and training details}\label{sec:impl}

Due to the nature of the therapy dataset, we have specified the memory length $w$ of our TCN to be 32 frames long by using a convolutional kernel size of 4, stacks of 2 residual blocks, and dilations of 1, 2 and 4  for the layers within  each convolutional block. We also use skip connections and 256 filters in each convolutional layer, which finally generates motion descriptors of size 256.

This TCN is pretrained for a classification task (categorical cross-entropy loss) in the NTU RGB+D 120 dataset by adding an output classification layer to the TCN backbone and applying the following random data augmentation to the original coordinates:

\begin{itemize}
    \item \textbf{Movement speed variation}. Joint coordinates are randomly scaled by interpolation along the time dimension to simulate varying movement speeds.
    \item \textbf{Skip frames}. Since the training dataset is recorded at more than twice the frame rate of the evaluation dataset, for each training sample we just use one out of either two or three frames, discarding the rest.
    \item \textbf{Horizontal flip}. Coordinates that correspond to the left or right part of the body are randomly flipped. Vertical and depth dimensions remain as originally. 
    \item \textbf{Random cropping}. Actions longer than the receptive field are randomly cropped to fit the TCN memory length. Shorter ones are pre-padded with zeros.
\end{itemize}

Finally, after calculating and stacking all the pose features $M$ the input training data has a size of $32 \times 423$. 

After training, the output classification layer is discarded to extract the 256-dimensional motion descriptors straight from the TCN backbone.

\subsection{Validation of the system on a generic one-shot action recognition benchmark}\label{sec:eval_ntu}

Table \ref{tab:ntu_one_shot} shows the classification accuracy of our framework for one-shot action recognition on the NTU-120 one-shot 3D action recognition problem, where we outperform the results of previous action recognition models \cite{liu2019ntu}. For this benchmark, we use the same implementation described in Section \ref{sec:impl} and the same training procedure,
but with the train/test splits specified in the original paper \cite{liu2019ntu}. 
Additionally, we also set the frame skipping to 2 both for training and evaluation and we suppress the random horizontal flipping during training. 
Descriptors are evaluated with the cosine distance, but other distance functions such as the Jensen-Shannon divergence report identical or comparable performance.
Since this problem is about classifying action segments and not identifying anchor actions in the wild, classification is performed by assigning, for each evaluation sample, the class with the lowest similarity distance among the set of anchor action segments.
Note that, unlike the other action recognition models that work in an offline fashion, our model works in an online and real-time fashion and only uses half of the available data (alternate frames).

\begin{table}[!tb]
\centering
\small
\begin{tabular}{|l|c|}
\hline
\textbf{Method}                    & \textbf{Accuracy} \\ \hline
ST-LSTM \cite{liu2017skeleton} + Average Pooling & 42.9\%              \\ \hline
ST-LSTM \cite{liu2017skeleton} + Fully Connected & 42.1\%              \\ \hline
ST-LSTM \cite{liu2017skeleton} + Attention       & 41.0\%              \\ \hline
APSR\cite{liu2019ntu}                      & 45.3\%              \\ \hline
\textbf{Ours} (one-shot, m=1)             & \textbf{46.5} \%    \\ \hline
\end{tabular}
\caption{NTU RGB+D 120 Dataset One-shot evaluation}
\label{tab:ntu_one_shot}
\end{table}

\subsection{Validation and discussion on real therapies data}

This experiment evaluates the performance of our action recognition approach on the imitation games from the therapy dataset. Here, the actions from the therapist are taken as anchor to later detect their imitation at each time-step of the patient motion stream. 

\subsubsection{Quantitative results}\label{sec:val_ther_quant}
\paragraph{Metrics.}
Action recognition is evaluated with common metrics, i.e. precision, recall and F1. However, since each target motion descriptor $z_{T}(n)$ does not refer only to a single time-step $n$ but to the motion from the $w$ previous frames, we have defined the following terms as follows:

\begin{itemize}
    \item \textbf{True Positive (TP)}. An action is correctly detected when the ground truth action is being executed or has been recently executed within the TCN receptive field $w$. A groundtruth action referenced by many detections only counts as one TP.
    \item \textbf{False Positive (FP)}. An action is incorrectly detected when no groundtruth action has been recently executed. Consecutive detections inside the TCN receptive field $w$ only count as one FP.
    \item \textbf{False Negative (FN)}. A groundtruth action is missed if it has not been referenced by any action detection. Each miss-detected action counts as a single FN.
\end{itemize}

\begin{figure}[!tb]
\centering
\includegraphics[width=0.99\linewidth]{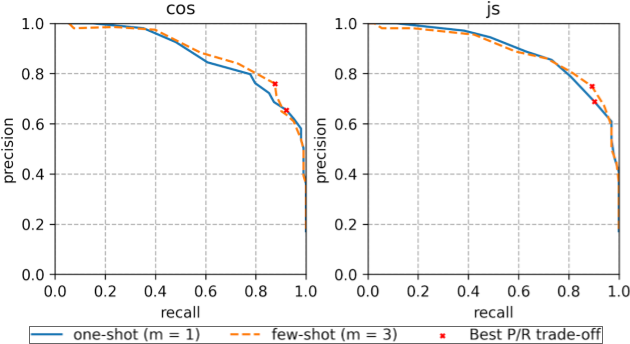}
\caption{
Precision/recall curves for the two evaluation distances used: cosine distance (cos) and Jensen-Shannon divergence (js) . Solid lines refer to regular one-shot evaluations ($m=1$) and dashed lines refer to few-shot recognition with extended anchor representations ($m=3$). Red crosses refer to the optimal F1 values.
}
\label{fig:pr_curve}
\end{figure}

\paragraph{Precision-Recall trade off.}
The threshold $\alpha$ selected to be used over the distance scores to perform action recognition is the one that optimizes the trade-off between precision and recall defined by the F1 metric. 
In our experiments, we calculate different thresholds to optimize the results achieved when using one descriptor per anchor ($m=1$) and with extended anchor representations ($m=3$). 
As seen in the precision-recall curves from Fig. \ref{fig:pr_curve}, the best trade-off is achieved at the cost of lowering the precision of the framework, especially when working with single anchor descriptors.

\begin{table*}[!tb]
\centering
\small
\begin{tabular}{|c|c|c|c|c|c|c|c|c|c|c|}
\hline
\multirow{2}{*}{\textbf{\begin{tabular}[c]{@{}c@{}}Recognition\\ Modality\end{tabular}}} & \multirow{2}{*}{\textbf{\begin{tabular}[c]{@{}c@{}}m \end{tabular}}} & \multirow{2}{*}{\textbf{\begin{tabular}[c]{@{}c@{}}Dynamic\\ Threshold\end{tabular}}} & \multicolumn{4}{l|}{\textbf{Cosine distance} } & \multicolumn{4}{l|}{\textbf{Jensen-Shannon divergence} } \\ \cline{4-11} 
 &  &  & \textbf{$\alpha$} & \textbf{Precision} & \textbf{Recall} & \textbf{F1} & \textbf{$\alpha$} & \textbf{Precision} & \textbf{Recall} & \textbf{F1} \\ \Xhline{4\arrayrulewidth}
one-shot & 1 &  & 0.48 & 0.655 & 0.922 & 0.697 & 0.50 & 0.689 & 0.902 & 0.714 \\ \hline
one-shot & 3 &  & 0.40 & 0.773 & 0.837 & 0.724 & 0.48 & 0.765 & 0.853 & 0.728 \\ \hline
one-shot & 3 & \checkmark & $\le$ 0.40 & 0.803 & 0.837 & \textbf{0.751} & $\le$ 0.48 & 0.779 & 0.853 & \textbf{0.739} \\ 
\Xhline{4\arrayrulewidth}
few-shot* & 1 &  & 0.48 & 0.784 & 0.827 & 0.731 & 0.50 & 0.785 & 0.846 & 0.744 \\ \hline %\Xhline{4\arrayrulewidth}
% few-shot* & 1 &  & 0.40 & 0.784 & 0.827 & 0.731 & 0.48 & 0.785 & 0.846 & 0.744 \\ \hline
few-shot* & 3 &  & 0.40 & 0.760 & 0.876 & 0.753 & 0.48 & 0.749 & 0.892 & 0.755 \\ \hline
few-shot* & 3 & \checkmark & $\le$ 0.40 & 0.790 & 0.876 & \textbf{0.781} & $\le$ 0.48 & 0.763 & 0.892 & \textbf{0.765} \\ \hline
\multicolumn{11}{l}{\footnotesize * few-shot stands for the use of 1, 2 or 3 anchor sequences depending on the available reference data for each test, as explained in Section \ref{sec:datasets}}
\end{tabular}
\caption{Comparison of the studied variations of our action recognition approach on the evaluation therapy dataset. 
}
\label{tab:ablation}
\end{table*}

\paragraph{Analysis of variations of our approach.}
The effect of the different variations proposed to make our action recognition more robust are summarized in Table \ref{tab:ablation}. First three rows show the performance of the one-shot version, i.e., using as anchor the last action performed by the therapist in each imitation game. 
The use of extended anchor representations ($m=3$) allows us to have a more restrictive threshold $\alpha$, which reduces false positives, increasing the detection precision. 
The increase in precision is also achieved by making the threshold dynamic. New threshold values can be set differently in each imitation game by evaluating the anchor representation and the target motion up the end of the anchor execution (period in which the patient is not performing any specific action). The dynamic threshold avoids false positives related to "static" anchor actions, at no recall cost.
Finally, as expected, the performance improves when more anchor actions (up to 3 in the experiments) are available to run a few-shot recognition.
Regarding the embedding similarity computation, both the cosine distance and the JS divergence report similar performance when using extended anchor representations. However, the JS divergence performs better in simpler set-ups and the cosine distance excels when using a dynamic threshold.

The results can be further scrutinized in Table \ref{tab:f1_per_class}, that compares the recognition performance by action category. 
Extended anchor representations ($m=3$) and few-shot recognition significantly overcome the limitations of the simple one-shot recognition for challenging classes (last 8 rows of the table refer to actions with softer and subtler arm movements or actions that consist in staying on specific positions). This is further improved when using a dynamic threshold (DT).

\begin{table}[!t]
\centering
\small
\begin{tabular}{l|ccc}
\textbf{\makecell{Class\\name}}      & \textbf{\makecell{One-shot\\(m=1)}} & \textbf{\makecell{Few-shot\\(m=3)}} & \textbf{\makecell{Few-shot+DT\\(m=3)}} \\
\hline
\textbf{big}        & 1.00             & 1.00 (+0.00)     & 1.00 (+0.00)             \\
\textbf{high}       & 1.00             & 1.00 (+0.00)     & 1.00 (+0.00)             \\
\textbf{happy}      & 0.96             & 0.96 (+0.00)     & 0.96 (+0.00)             \\
\textbf{waving}     & 0.89             & 0.95 (+0.06)     & 0.95 (+0.06)             \\
\textbf{pointing}   & 0.89             & 0.89 (+0.00)     & 0.89 (+0.00)             \\
\textbf{giving}     & 0.76             & 0.76 (+0.00)     & 0.76 (+0.00)             \\
\textbf{small}      & 0.68             & 0.89 (+0.21)     & 0.96 (+0.28)             \\
\textbf{coming}     & 0.65             & 0.69 (+0.04)     & 0.69 (+0.04)             \\
\textbf{waiting}    & 0.63             & 0.73 (+0.10)     & 0.73 (+0.10)             \\
\textbf{hungry}     & 0.47             & 0.35 (-0.12)     & 0.42 (-0.05)             \\
\textbf{where}      & 0.47             & 0.61 (+0.14)     & 0.61 (+0.14)             \\
\textbf{down}       & 0.45             & 0.54 (+0.08)     & 0.54 (+0.08)             \\
\textbf{me}         & 0.44             & 0.67 (+0.22)     & 0.67 (+0.22)             \\
\textbf{angry}      & 0.31             & 0.30 (-0.01)     & 0.50 (+0.19)            
\end{tabular}
\caption{Per-class F1 evaluation comparison of different variations of our action recognition approach (using the cosine distance).
}
\label{tab:f1_per_class}
\end{table}

\begin{table}[!tb]
\centering
\small
\begin{tabular}{|c|c|c|c|}
\hline
\textbf{Coordinates} & \textbf{\begin{tabular}[c]{@{}c@{}}Normalized\\ Coordinates\end{tabular}} & \textbf{\begin{tabular}[c]{@{}c@{}}Geometric\\ Features\end{tabular}} & \textbf{F1} \\ \hline
\checkmark &  &  & 0.711 \\ \hline
 & \checkmark &  & 0.689 \\ \hline
 &  & \checkmark & 0.757 \\ \hline
\checkmark &  & \checkmark & 0.695 \\ \hline
 & \checkmark & \checkmark & \textbf{0.781} \\ \hline
\end{tabular}
\caption{F1 evaluation of different pose features running our top-performing action recognition (few-shot, cosine distance, extended anchor representations and dynamic threshold).}
\label{tab:feat-ablation}
\end{table}

\paragraph{Influence of different pose feature sets.}
Table \ref{tab:feat-ablation} reports the influence 
of different pose features proposed in Section \ref{sec:feat_set}.
Using only normalized coordinates 
achieves lower performance than using just the original body coordinates. Normalized coordinates gain in invariance, but they lose certain discriminative power.
The proposed geometric feature set achieves good performance on its own.
However, the best performance comes from the combination of normalized coordinates and geometric features, which suggests that the geometric features work better along with the invariance provided by normalized coordinates.

\subsubsection{Qualitative results}

\begin{figure}[tb]
\centering
\includegraphics[width=0.98\linewidth]{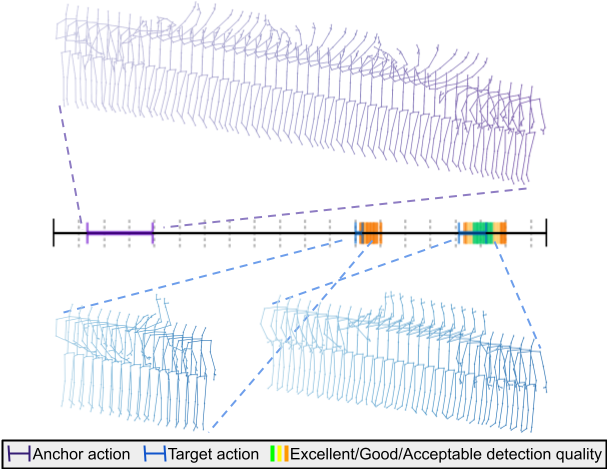}
\caption{
Action recognition on one imitation game from the therapy dataset. Central horizontal line represents the execution timeline, with one anchor action (top/purple) performed by the therapist, followed by two imitations (bottom/blue) performed by the patient.  
Dashed line segments indicate the groundtruth action time segmentations. Vertical small solid segments, ranging from green to orange, refer to time steps where the action has been detected, and the color gives an estimation of the repetition quality.
}
\label{fig:ex_sample}
\end{figure}

Figure \ref{fig:ex_sample} shows the timeline of an imitation game sampled from the therapy dataset. This exercise involves the action of raising and moving the arms, which is performed once (anchor) by the therapist and repeated twice (target) by the patient. Detection results from the timeline represent both the location and quality of both action repetitions. The latter is estimated according to the calculated similarity between anchor and target motion. The first action imitation is performed quickly (worse quality) and the second one slower and more detailed (better quality). For a better understanding, the whole execution of the experiment is shown in the supplementary video\footnote{https://sites.google.com/a/unizar.es/filovi/}.

\subsubsection{Time performance}

Temporal Convolutional Networks are frequently lightweight models, making them suitable for low-resource environments like in the therapies described in Section \ref{sec:datasets}.
With the settings described in Section \ref{sec:impl}, 
the action recognition (except the skeleton extraction from RGB-D data)
takes 0.08 ms per time-step using the GPU and 
0.1 ms
just using the CPU\footnote{Performing speed has been calculated with a NVIDIA GeForce RTX 2080 Ti (GPU) and a Intel Core i7-9700K (CPU)}. Time performance has been calculated with the therapy dataset skeleton sequences. Note that due to the TCN parallel computing, the longer the motion sequences are, the faster per-frame processing we can get.

%%%%%%%%%%%%%%%%%%%%%%%%%%%%%%%%%%%%%%%%%%%%%%%%%%%%%%%%%%%%%%%%%%%%%%%%%%%%%%%%
\section{Conclusions}

This work presents a novel skeleton-based framework for one-shot action recognition in the wild. 
Our method generates motion descriptors 
robust different motion artifacts and variable kinematics. We achieve accurate action recognition by combining the proposed set of pose features and an efficient architecture based on a TCN that encodes these pose features. 
Besides, we demonstrate the effectiveness of several simple steps included in our method to boost the action recognition performance in challenging real world scenarios, i.e. with limited reference data and noisy repetitions of the actions to be recognized.
Our base one-shot recognition approach is evaluated on the public NTU RGB+D 120 dataset, outperforming previous action recognition methods, by using only half of the available data. 
We also demonstrate the suitability of our framework to analyze videos from real therapies with autistic people. These recordings are characterized by having extreme motion artifacts.
Evaluation results provide both quantitative and qualitative measures about the actions recognized, which is essential for the patient evaluation and monitoring. Moreover, our approach can run online, being able to provide immediate feedback to the therapist and making the sessions more dynamic.

\section*{Acknowledgments}
This research has been funded by FEDER/Ministerio de Ciencia, Innovación y Universidades/Agencia Estatal de Investigación RTC-2017-6421-7
and PGC2018-098817-A-I00, DGA T45 17R/FSE, the Office of Naval Research Global project ONRG-NICOP-N62909-19-1-2027, Universidad de Zaragoza, Fundación Bancaria Ibercaja and Fundación CAI IT 17/19, and the Funda\c{c}\~ao para a Ci\^encia e Tecnologia (FCT) project UIDB/50009/2020 and PhD scholarship SFRH/BD/145040/2019.

% {
% \bibliographystyle{IEEEtran}
% \bibliography{biblio}
% }
{\small
\bibliographystyle{ieee_fullname}
\bibliography{biblio}

\begin{thebibliography}{10}\itemsep=-1pt

\bibitem{bai2018empirical}
Shaojie Bai, J~Zico Kolter, and Vladlen Koltun.
\newblock An empirical evaluation of generic convolutional and recurrent
  networks for sequence modeling.
\newblock {\em Universal Language Model Fine-tuning for Text Classification},
  2018.

\bibitem{chen2010learning}
Cheng Chen, Yueting Zhuang, Feiping Nie, Yi Yang, Fei Wu, and Jun Xiao.
\newblock Learning a 3d human pose distance metric from geometric pose
  descriptor.
\newblock {\em IEEE Transactions on Visualization and Computer Graphics},
  17(11):1676--1689, 2010.

\bibitem{cheng2020skeleton}
Ke Cheng, Yifan Zhang, Xiangyu He, Weihan Chen, Jian Cheng, and Hanqing Lu.
\newblock Skeleton-based action recognition with shift graph convolutional
  network.
\newblock In {\em Proceedings of the IEEE/CVF Conference on Computer Vision and
  Pattern Recognition}, pages 183--192, 2020.

\bibitem{donahue2015long}
Jeffrey Donahue, Lisa Anne~Hendricks, Sergio Guadarrama, Marcus Rohrbach,
  Subhashini Venugopalan, Kate Saenko, and Trevor Darrell.
\newblock Long-term recurrent convolutional networks for visual recognition and
  description.
\newblock In {\em Proceedings of the IEEE conference on computer vision and
  pattern recognition}, pages 2625--2634, 2015.

\bibitem{fanello2013one}
Sean~Ryan Fanello, Ilaria Gori, Giorgio Metta, and Francesca Odone.
\newblock One-shot learning for real-time action recognition.
\newblock In {\em Iberian Conference on Pattern Recognition and Image
  Analysis}, pages 31--40. Springer, 2013.

\bibitem{hahn2019action2vec}
Meera Hahn, Andrew Silva, and James~M Rehg.
\newblock Action2vec: A crossmodal embedding approach to action learning.
\newblock 2019.

\bibitem{jasani2019skeleton}
Bhavan Jasani and Afshaan Mazagonwalla.
\newblock Skeleton based zero shot action recognition in joint pose-language
  semantic space.
\newblock {\em arXiv preprint arXiv:1911.11344}, 2019.

\bibitem{kim2017interpretable}
Tae~Soo Kim and Austin Reiter.
\newblock Interpretable 3d human action analysis with temporal convolutional
  networks.
\newblock In {\em 2017 IEEE conference on computer vision and pattern
  recognition workshops (CVPRW)}, pages 1623--1631. IEEE, 2017.

\bibitem{konevcny2014one}
Jakub Kone{\v{c}}n{\`y} and Michal Hagara.
\newblock One-shot-learning gesture recognition using hog-hof features.
\newblock {\em The Journal of Machine Learning Research}, 15(1):2513--2532,
  2014.

\bibitem{liu2019ntu}
Jun Liu, Amir Shahroudy, Mauricio~Lisboa Perez, Gang Wang, Ling-Yu Duan, and
  Alex~Kot Chichung.
\newblock Ntu rgb+ d 120: A large-scale benchmark for 3d human activity
  understanding.
\newblock {\em IEEE transactions on pattern analysis and machine intelligence},
  2019.

\bibitem{liu2017skeleton}
Jun Liu, Amir Shahroudy, Dong Xu, Alex~C Kot, and Gang Wang.
\newblock Skeleton-based action recognition using spatio-temporal lstm network
  with trust gates.
\newblock {\em IEEE transactions on pattern analysis and machine intelligence},
  40(12):3007--3021, 2017.

\bibitem{liu2017global}
Jun Liu, Gang Wang, Ping Hu, Ling-Yu Duan, and Alex~C Kot.
\newblock Global context-aware attention lstm networks for 3d action
  recognition.
\newblock In {\em Proceedings of the IEEE Conference on Computer Vision and
  Pattern Recognition}, pages 1647--1656, 2017.

\bibitem{liu2020disentangling}
Ziyu Liu, Hongwen Zhang, Zhenghao Chen, Zhiyong Wang, and Wanli Ouyang.
\newblock Disentangling and unifying graph convolutions for skeleton-based
  action recognition.
\newblock In {\em Proceedings of the IEEE/CVF Conference on Computer Vision and
  Pattern Recognition}, pages 143--152, 2020.

\bibitem{mikolov2013distributed}
Tomas Mikolov, Ilya Sutskever, Kai Chen, Greg~S Corrado, and Jeff Dean.
\newblock Distributed representations of words and phrases and their
  compositionality.
\newblock In C.~J.~C. Burges, L. Bottou, M. Welling, Z. Ghahramani, and K.~Q.
  Weinberger, editors, {\em Advances in Neural Information Processing Systems},
  volume~26. Curran Associates, Inc., 2013.

\bibitem{oord2016wavenet}
Aaron van~den Oord, Sander Dieleman, Heiga Zen, Karen Simonyan, Oriol Vinyals,
  Alex Graves, Nal Kalchbrenner, Andrew Senior, and Koray Kavukcuoglu.
\newblock Wavenet: A generative model for raw audio.
\newblock {\em arXiv preprint arXiv:1609.03499}, 2016.

\bibitem{pagliardini2017unsupervised}
Matteo Pagliardini, Prakhar Gupta, and Martin Jaggi.
\newblock Unsupervised learning of sentence embeddings using compositional
  n-gram features.
\newblock In {\em Proceedings of the 2018 Conference of the North American
  Chapter of the Association for Computational Linguistics: Human Language
  Technologies, Volume 1 (Long Papers)}, pages 528--540, 2018.

\bibitem{perez2021interaction}
Mauricio Perez, Jun Liu, and Alex~C Kot.
\newblock Interaction relational network for mutual action recognition.
\newblock {\em IEEE Transactions on Multimedia}, 2021.

\bibitem{santos2020interactive}
Laura Santos, Silvia Annunziata, Alice Geminiani, Elena Brazzoli, Arianna
  Caglio, Jos{\'e} Santos-Victor, Alessandra Pedrocchi, and Ivana Olivieri.
\newblock Interactive social games with a social robot ({IOGIOCO}):
  Communicative gestures training for preschooler children with autism spectrum
  disorder.
\newblock In {\em I {Congresso} {Annuale} {Rete} {IRCCCS} {Neuroscienze} e
  {Neuroriabilitazione}}, 2020.

\bibitem{Santos2019}
Laura Santos, Alice Geminiani, Ivana Olivieri, Jos{\'e} Santos-Victor, and
  Alessandra Pedrocchi.
\newblock Copyrobot: Interactive mirroring robotics game for asd children.
\newblock In Jorge Henriques, Nuno Neves, and Paulo de Carvalho, editors, {\em
  XV Mediterranean Conference on Medical and Biological Engineering and
  Computing -- MEDICON 2019}, pages 2014--2027, Cham, 2020. Springer
  International Publishing.

\bibitem{su2020predict}
Kun Su, Xiulong Liu, and Eli Shlizerman.
\newblock Predict \& cluster: Unsupervised skeleton based action recognition.
\newblock In {\em Proceedings of the IEEE/CVF Conference on Computer Vision and
  Pattern Recognition}, pages 9631--9640, 2020.

\bibitem{vinyals2016matching}
Oriol Vinyals, Charles Blundell, Timothy Lillicrap, Daan Wierstra, et~al.
\newblock Matching networks for one shot learning.
\newblock In {\em Advances in neural information processing systems}, pages
  3630--3638, 2016.

\bibitem{yang2019make}
Fan Yang, Yang Wu, Sakriani Sakti, and Satoshi Nakamura.
\newblock Make skeleton-based action recognition model smaller, faster and
  better.
\newblock In {\em Proceedings of the ACM Multimedia Asia}, pages 1--6. 2019.

\bibitem{zhang2019view}
Pengfei Zhang, Cuiling Lan, Junliang Xing, Wenjun Zeng, Jianru Xue, and Nanning
  Zheng.
\newblock View adaptive neural networks for high performance skeleton-based
  human action recognition.
\newblock {\em IEEE transactions on pattern analysis and machine intelligence},
  41(8):1963--1978, 2019.

\bibitem{zhang2017geometric}
Songyang Zhang, Xiaoming Liu, and Jun Xiao.
\newblock On geometric features for skeleton-based action recognition using
  multilayer lstm networks.
\newblock In {\em 2017 IEEE Winter Conference on Applications of Computer
  Vision (WACV)}, pages 148--157. IEEE, 2017.

\bibitem{zhang2012spatio}
Yimeng Zhang, Xiaoming Liu, Ming-Ching Chang, Weina Ge, and Tsuhan Chen.
\newblock Spatio-temporal phrases for activity recognition.
\newblock In {\em European Conference on Computer Vision}, pages 707--721.
  Springer, 2012.

\bibitem{zhang2012microsoft}
Zhengyou Zhang.
\newblock Microsoft kinect sensor and its effect.
\newblock {\em IEEE multimedia}, 19(2):4--10, 2012.

\end{thebibliography}
}

\end{document}